\documentclass[utf8]{FrontiersinHarvard} 

\usepackage{url,hyperref,lineno,microtype,subcaption}
\usepackage[onehalfspacing]{setspace}
\def\@mb@citenamelist{cite,citep,citet,citealp,citealt,citepalias,citetalias}
\makeatother
\usepackage{CJKutf8}

\def\keyFont{\fontsize{8}{11}\helveticabold }
\def\firstAuthorLast{Xiao {et~al.}} 
\def\Authors{Yunze Xiao\,$^{1}$, Houda Bouamor\,$^{1*}$ and Wajdi Zaghouani \,$^{2*}$}
\def\Address{$^{1}$Carnegie Mellon University Qatar,Doha.Qatar \\
$^{2}$Hamad Bin Khalifa University}
\def\corrAuthor{Houda Bouamor, Wajdi Zaghouani}

\def\corrEmail{hbouamor@andrew.cmu.edu,wzaghouani@hbku.edu.qa}

\begin{document}
\begin{CJK*}{UTF8}{gbsn}
\onecolumn
\firstpage{1}

\title[]{\textbf{Chinese Offensive Language Detection:\\Current Status and Future Directions}} 

\author[\firstAuthorLast ]{\Authors} 
\address{} 
\correspondance{} 

\extraAuth{}

\maketitle

\begin{abstract}

\section{}
Despite the considerable efforts being made to monitor and regulate user-generated content on social media platforms, the pervasiveness of offensive language, such as hate speech or cyberbullying,  in the digital space remains a significant challenge. Given the importance of maintaining a civilized and respectful online environment, there is an urgent and growing need for automatic systems capable of detecting offensive speech in real time. However, developing effective systems for processing languages such as Chinese presents a significant challenge, owing to the language's complex and nuanced nature, which makes it difficult to process automatically. This paper provides a comprehensive overview of offensive language detection in Chinese, examining current benchmarks and approaches and highlighting specific models and tools for addressing the unique challenges of detecting offensive language in this complex language. The primary objective of this survey is to explore the existing techniques and identify potential avenues for further research that can address the cultural and linguistic complexities of Chinese.

\tiny
 \keyFont{ \section{Keywords:} Chinese Offensive Language, Chinese Benchmarks, Toxic Content Online,Chinese social media,Cultural and linguistic complexities} 
\end{abstract}

\section{Introduction}
\textbf{Disclaimer: This paper contains some examples of hateful content. This is strictly for the purposes of enabling this research, and we have sought to minimize the number of examples where possible. Please be aware that this content could be offensive and cause you distress.}

Hate speech detection represents a critical frontier in the fight against online toxicity. Detecting offensive language, in general, is a crucial task for maintaining a civilized, safe, and respectful online environment and ensuring the well-being of users in the digital space \cite{deng-etal-2022-cold}. With the rapid growth and increasing use of social media platforms, offensive language has become pervasive, taking many forms, such as hate speech, cyberbullying, and adult-oriented content \cite{2001.03131}. Through machine learning and linguistic analysis, the goal has always been: to build robust solutions that protect users from hate speech and its damaging impact. Offensive language detection has been studied for various languages, such as English~\cite{davidson2017automated,pitsilis2018effective,wei2021offensive}, Arabic~\cite{fatemah2021survey}, French~\cite{battistelli2020building}, Turkish~\cite{beyhan2022turkish} with a recent interest in Asian languages~\cite{dhanya2021hate} and Chinese~\cite{deng-etal-2022-cold,2303.17927}. But, despite the increasing interest in detecting offensive language in Chinese, many studies fail to fully acknowledge the unique challenges posed by the language and culture. The Chinese language's contextual-based nature, its wide range of dialects and regional variations, and the prevalence of subversive expressions make it difficult for automatic systems to detect offensive language accurately. Moreover, Chinese is heavily influenced by various cultural references, adding further complexity to the task \cite{2303.17927}. For instance, the expression "ham ga fo gui"(咸家富贵/全家富贵) is used to mean "may your whole family die" in Cantonese-speaking areas and "may your whole family get rich" in Mandarin context. The expression "grass mud horse"(草泥马) is an example of subversive language, which is used to represent the highly offensive phrase "f**k your mother"(操你妈). Furthermore, A Chinese character appearing in an offensive word may also appear in an inoffensive one. For example, although the character "干" means "fxxk" in some contexts, it also has meanings of "do" and "work", as its appearances in the words “干活” (do works)~\cite{su2017rephrasing}. These kinds of phrases are not easy to detect by current offensive language detection systems underscoring the necessity for researchers to conduct in-depth investigations into their recognition and handling to develop more effective systems for detecting offensive language in Chinese.

In this paper, we present a comprehensive survey that reviews prior research on offensive language detection in the Chinese language. We summarize the literature covering state-of-the-art benchmarks and techniques, with a focus on identifying open challenges and research gaps. By addressing these issues and highlighting opportunities for further investigation, our objective is to advance the development of effective offensive language detection systems for Chinese. Our work involves a detailed examination of existing benchmarks, approaches, and state-of-the-art models, emphasizing the unique focus on the Chinese language. We explore cross-cultural transfer learning, translational techniques, and advanced models while also identifying intriguing research ideas and challenges, including subversive expression disambiguation, the incorporation of cultural context~\cite{2001.03131}, and data labeling concerns. The importance and relevance of detecting offensive language in Chinese lie in the fact that it is one of the most widely spoken languages in the world, and understanding the nuances of offensive language in Chinese and its variants can contribute to the development of more effective and culturally sensitive systems \cite{1901.05816}. 

The paper is structured as follows. Section ~\ref{sec:background} defines offensive language and introduces the background and unique challenges that occur in the Chinese language.  Section~\ref{sec:benchmarks} presents the current benchmarks used for offensive language detection in Chinese. Section~\ref{section:approaches} explores the state-of-the-art models and approaches developed for this task. Section~\ref{sec:potential} highlights existing research gaps. Section~\ref{sec:challenges} presents the challenges that researchers in this task. The conclusion is given in Section~\ref{sec:conclusion}.

\section{Background}
\label{sec:background}
Identifying offensive text is a complex task, as it relies on subjective interpretations and the definition of offensiveness. Hence, we start by providing definitions of offensive language in the context of Chinese and categorizing the common types of offensive expressions. We also discuss the specific challenges posed by the Chinese language.

\subsection{Offensive Language}
Offensive language refers to words or expressions that are considered inappropriate, insulting, or disrespectful towards a particular individual or group, often on the basis of their race, ethnicity, gender, religion, sexual orientation, or other personal characteristics~\cite{jay2008pragmatics,davidson2019racial}. Such language may include slurs, insults, profanity, or other forms of derogatory language. The use of offensive language can have a harmful impact on the targeted individuals or groups, contributing to discrimination, prejudice, and hate speech.

\paragraph*{Hate Speech} is defined as a speech targeted at a specific group of people with the intention of causing harm, violence, or social disruption~\cite{waseem-etal-2017-understanding,sigurbergsson-derczynski-2020-offensive,ajvazi-hardmeier-2022-dataset}, regardless of the language used. Common examples of hate speech include the use of racial and homophobic slurs. \citeauthor{waseem-etal-2017-understanding} established a list of rules to identify hate speech, including negative statements or stereotypes about minority groups, disparaging or racial terms, and support for entities that encourage hate speech. Hate speech is usually
done based on characteristics such as race, color, ethnicity, gender, sexual orientation, nationality, or religion. For instance, in Chinese, the term "婚驴" /the married donkey/ is commonly used, particularly by radical feminists to describe married women, which can be seen as a form of derogatory language targeting married women.

\paragraph*{Adult Content}
Adult content, intended for 18 and older audience, encompasses various materials, from sexual content to violence and explicit language~\cite{mubarak-etal-2021-adult}. Not all adult content is inherently offensive; its offensiveness often depends on context and cultural norms. In the Chinese context, adult content is conveyed implicitly due to strict regulations.  This has led to the widespread use of subversive phrases in offensive language~\cite{zidani-2018-represented}. For example, terms like “萝卜”(carrot)  are used to reference male genitals, and “奶子”(milk) is employed to refer to female breasts. These metaphorical expressions pose several challenges to the automatic detection of adult content.

\paragraph*{Sarcasm and Dark Humor}
Sarcasm and dark humor are two linguistic phenomena that have been studied extensively in linguistics~\cite{skalicky2018linguistic}, psychology~\cite{kreuz1989sarcastic}, and NLP~\cite{joshi2018understanding}. While these forms of speech can serve as sources of entertainment in non-offensive contexts, they are more frequently encountered in potentially offensive situations. Sarcasm presents a controversial aspect of offensive language, largely due to its varying interpretations. In this paper, we adopt the definition from merriam-webster: "\textit{a sharp and often satirical or ironic utterance designed to cut or give pain}".\footnote{\url{https://www.merriam-webster.com/dictionary/sarcasm}} Using this definition, we categorize sarcasm that fits this description as offensive language, given its intent to cause harm. While previous research hasn't explicitly included sarcasm as a type of offensive language, several studies link sarcasm to negative sentiments like anger or jealousy \cite{Riloff2013SarcasmAC,abu-farha-magdy-2020-arabic}. Failing to classify sarcasm as offensive language can overlook its potential harm \cite{Frenda2022KillingMS,Frenda2022TheUH}. Hence, we include sarcasm as a form of offensiveness. 

Conversely, dark humor entails employing humor to address taboo or morbid subjects, as noted in a study by \citet{mittal-etal-2021-think} . However, dark humor can cross into offensiveness when it turns cruel or targets marginalized or vulnerable groups, including those with disabilities, specific genders, races, sexualities, religions, or cultural identities~\cite{kumari-etal-2023-persuasive}. Social media platforms frequently feature dark humor, with memes incorporating such humor being widely shared~\cite{kiela-2020-hateful}.

Dark humor in Chinese can come in different forms, majority of them come from a historical perspective. Prominent ones such as "headless Louis XVI" and "farming equipment" (which refers to slavery in the 17th century) are used in a variety of contexts. However, this dark humor is heavily influenced by Western culture and has only become pervasive within a few years. Sarcasm, on the other hand, is more pervasive than dark humor in a Chinese context. For example, comments in Tieba often allude to the overweighted female as "健达奇趣蛋"(kinder egg).

\subsection{Chinese Language: Characteristics and Challenges}

Approximately 1.3 billion people around the world are using Chinese as their primary language. The Chinese language can be divided into many dialects, the most common being Mandarin, which is the official language of China and Singapore, and Cantonese, which is spoken in Singapore, Hong Kong, and Guangdong Province. Other dialects include 
Shanghainese, Hokkien, and Hakka, among others. The dialectal and regional variations coupled with the highly context- and culture-based nature of the language have
made several NLP tasks specifically challenging. Offensive content detection in Chinese presents several unique challenges. For instance, the Chinese language includes a vast number of characters, which can be combined to form new words and phrases. This leads to a high degree of flexibility in constructing hate speech, making it difficult to build comprehensive detection systems. Also, hate speech can be deeply embedded in cultural and historical contexts. Understanding the cultural nuances is crucial for accurate detection. In addition to text, online content in China often includes images, videos, and audio. Effective hate speech detection systems need to encompass these different modalities. To address these challenges, hate speech detection in the Chinese language requires sophisticated NLP models, cultural understanding, and continuous adaptation to the evolving linguistic landscape.

\section{Datasets for Chinese Offensive Language}
\label{sec:benchmarks}

In order to effectively address the issue of offensive language detection, it is essential to establish a dependable and adaptable benchmark as the foundational cornerstone for advancing in-depth research in this field. Quality datasets are vital for training machine learning models to detect offensive language in Chinese, as they provide labeled data for model training and evaluation.  These datasets are usually collected from various sources like social media, comments, and forums, forming the foundation for accurate and effective offensive language detection systems, given the language complexity. In this section, we present a few key datasets and benchmarks for this task, in Chinese.

\subsection{General Offensive Language Benchmarks}

\paragraph*{The Chinese Offensive Language Dataset (COLD)} is among the first benchmarks designed to facilitate Chinese offensive language detection and evaluation~\cite{deng-etal-2022-beike}. It includes a dataset and a baseline detector that are used to explore the task of offensive language detection in Chinese. The dataset contains 37,480 sentences labeled with four main categories: attacking individuals, attacking groups, anti-bias, and other non-offensive. In addition to addressing the challenge of Chinese offensive language detection, COLD investigates the factors that influence offensive generations and finds that anti-bias content and keywords referring to certain groups or revealing negative attitudes trigger offensive outputs more easily. Table~\ref{table:benchmarks} summarizes the state-of-the-art results for this task.

\begin{table}
\centering
\setlength{\tabcolsep}{2pt}
\begin{tabular}{l|c|c|c}
\hline
\textbf{Model} & \textbf{Train Set} & \textbf{Test F1} & \textbf{Test ACC} \\
\hline
XLM-R$_{\text{large}}^*$ & COLD+KOLD & 0.798 & 0.825 \\
COLDET & COLD & 0.810 & 0.810 \\
TJIGDET & Jigsaw dataset & 0.620 & 0.600 \\
BAIDUTC & Unknown & 0.540 & 0.630 \\
PSELFDET & BERT & 0.580 & 0.590 \\
\hline
\end{tabular}
\caption{State-of-the-art offensive language detection system's performance using the COLD benchmark}
\label{table:benchmarks}
\end{table}

Even though this dataset, was the first of its kind, it suffers from a lack of diversity as it only covers the topics of race, gender, and region and it does not adequately represent the full spectrum of offensive language in Chinese. Also, labeling offensive language can be highly subjective, and it may vary based on cultural context and personal interpretation. Furthermore, as offensive language evolves over time, the COLD dataset may not reflect the current landscape of offensive language on the internet.

\paragraph*{TOCP and TOCAB} TOCP~\cite{yang-lin-2020-tocp} is a dataset of Chinese profanity created for the purpose of detection and rephrasing. The dataset includes over 16,000 sentences from social media, with over 17,000 profane
expressions, extracted from PTT, a famous BBS site in Taiwan.\footnote{\url{https://www.ptt.cc/bbs/index.html}} A system that takes a profane expression and its preceding word as input and generates its possible reformulations was built using this dataset. In 2021, TOCAB\cite{9598528} was introduced as an extension to TOCP. The TOCAB dataset contains 1,000 posts collected from PTT, with 121,344 comment sentences, 17,836 of which are labeled as abusive and classified into six pre-defined abusive classes. Table~\ref{table:TOCAP} summarizes the state-of-the-art abusive language detection systems built using this dataset.

\begin{table}
\centering
\setlength{\tabcolsep}{2pt}
\begin{tabular}{lccc}
\hline
\textbf{Model} & \textbf{Precision} & \textbf{Recall} & \textbf{Test ACC} \\
\hline
BERT & 0.886 & 0.885 & 0.886 \\
BiGRUx3 & 0.900 & 0.830 & 0.806 \\
SVM & 0.910 & 0.762 & 0.829 \\
\hline
\end{tabular}
\caption{Offensive language detection system's performance using the TOCAB benchmark}
\label{table:TOCAB}
\end{table}

While TOCP and TOCAB are valuable language resources, it's important to acknowledge that their exclusive sourcing from Taiwan presents limitations. These datasets may not fully represent offensive language usage in mainland China, where cultural differences can significantly impact language dynamics. Moreover, the prevalence of traditional Chinese text in these datasets may pose distinctive challenges when adapting and fine-tuning pre-trained models for broader applications in mainland China. Therefore, while these datasets provide a valuable foundation, their effectiveness in addressing major tasks in mainland China may be compromised by cultural differences and the unique encoding of text, warranting careful consideration when applying them in such contexts.

\paragraph*{Sina Weibo Sexism Review (SWSR)} dataset~\cite{jiang2021swsr}, the first Chinese dataset focused on sexism. It is a comprehensive dataset created to investigate the diverse behaviors, beliefs, and attitudes expressed toward women in Chinese social media. The dataset was created by collecting sexism-related posts and comments from Sina Weibo, a Chinese microblogging website, and then annotated with three labels: (i) sexism or non-sexism, (ii) sexism category, and (iii) target type.  SWSR is also accompanied by a large lexicon called SexHateLex, which is the first lexicon in Chinese made of abusive and gender-related terms. This dataset was used to train models for sexist speech detection and provides a benchmark for sexism detection in Chinese. It also highlights the difficulty of identifying sexist targets in posts and the potential over-dependence of models on sexist words, which underscores the importance of developing comprehensive
and diverse datasets for researching hate. Table~\ref{table:SWSR} summarizes the state-of-the-art results for this task.

While the SWSR dataset and the SexHateLex lexicon are valuable resources for researching online sexism in Chinese social media, they face challenges in comprehensively capturing the evolving nature of sexist speech. The reliance on specific keywords may limit their ability to identify nuanced expressions of sexism. Moreover, the efficacy of models trained on such datasets can be affected by the constantly changing dynamics of online discourse.

\begin{table}
\centering
\setlength{\tabcolsep}{2pt}
\begin{tabular}{lcc}
\hline
\textbf{Model} & \textbf{Test F1} & \textbf{Test ACC} \\
\hline
BERT & 0.776 & 0.806 \\
Bert-wwm & 0.762 & 0.792 \\
RoBerta & 0.764 & 0.792 \\
SVM + ngram & 0.739 & 0.739 \\
\hline
\end{tabular}
\caption{State-of-the-art offensive language detection system's performance using the SWSR benchmark}
\label{table:SWSR}
\end{table}

\paragraph*{Categorizing Offensive Language (COLA)} \cite{tang-shen-2020-categorizing}, introduced as the first Chinese offensive language classification dataset, is a large-scale dataset of offensive texts in Chinese built by crawling comments from YouTube and Weibo, resulting in a total of 18.7k comments. Three annotators categorized the texts into four classes: neutral, insulting, antisocial, and illegal. This dataset is the first of its kind and includes user-generated comments from different social media platforms. Using this dataset, four automatic classification systems were built and evaluated.

\subsection{Chinese Sarcasm Datasets}

There were several datasets introduced to detect sarcasm in Chinese.  Among them, the Chinese Sarcasm Dataset \cite{gong-etal-2020-design} is a large
and high-quality dataset of sarcastic and non-sarcastic texts in the Chinese language. The dataset contains 2,486 manually annotated sarcastic texts and 89,296 non-sarcastic texts. More recently, \citet{OpenSarcasm} introduced the balanced open Chinese Internet sarcasm corpus. The corpus consists of 2,000 texts that were selected from larger datasets labeled to indicate whether they are sarcastic or not. Similarly, \citet{zhang2022novel} introduced a dataset extracted from Bilibili, a Youtube-like website with a variety of sarcastic information on specific topics.\footnote{\url{https://www.bilibili.com/}} While these datasets were used to make significant strides in sarcasm detection and data collection from various online platforms, a critical aspect has been largely overlooked: the connection between sarcasm and offensiveness. Sarcasm is a complex form of communication that often relies on irony and humor, making its interpretation context-dependent. What may be intended as playful sarcasm in one setting can easily be perceived as offensive in another.


\subsection{Toxic Content Dataset}
In the realm of Chinese offensive language detection, the TOXICN dataset represents a significant milestone~\cite{lu-etal-2023-facilitating}. This dataset was meticulously constructed by collecting data from prominent online platforms such as Zhihu and Tieba, followed by a series of filtering and labeling measures.\footnote{\url{https://www.zhihu.com/}, \url{http://c.tieba.baidu.com/}} The resulting dataset features a sophisticated multi-level labeling system, encompassing categories like offensive language, hate speech, targeted groups, and expression types. This comprehensive approach offers researchers valuable insights into the complexities and nuances of offensive language in the Chinese context.
TOXICN was used to introduce a benchmark known as "Toxic Knowledge Enhancement (TKE)" designed to enhance the detection of toxic language using lexical knowledge. TKE involves enriching the representation of sentences by incorporating lexical features. This is achieved by embedding each token of a sentence in a vector space and introducing toxic knowledge through a specific embedding, which categorizes tokens as related to insults. The enhanced token representation combines the toxic embedding with the original word embedding, enabling a more effective integration of toxic and linguistic information. TKE can be applied to any pre-trained large language models to enhance the detection of toxic language.

The TOXICN dataset undeniably marks a significant advancement in the field of Chinese offensive language detection. However, it's essential to critically examine certain aspects of it. One notable concern is that the percentage in offensive language in the platform tieba is significantly higher than other platform. While the dataset creators may have had valid reasons for this labeling, such  categorization raises questions about potential biases and the discriminatory effects it might have on the platform's users. Additionally, while the TKE benchmark proposes an intriguing approach to enhancing toxic language detection, its effectiveness across various offensive language contexts and its potential biases should be subject to rigorous evaluation to ensure it aligns with the goal of reducing harm without unjustly censoring legitimate discourse.

\section{Current Approaches and Models}
\label{section:approaches}

In this section, we present an overview of the state-of-the-art approaches and models introduced to tackle the task of offensive language detection in Chinese.

\paragraph*{Lexicon-based Models} These models have proven to be effective in offensive language detection~\cite{Zhang2010UnderstandingBM}. One common technique employed in this field is keyword matching~\cite{deng-etal-2022-beike} using a predefined lexicon. \citet{su2020chinese}, for instance, presented a method that leverages BERT and LDA for identifying abusive language in Chinese text. However, it had limitations, including the fact that a predefined lexicon may not encompass all potentially offensive terms, and LDA modeling may fail to capture the nuanced nature of offensive language. \citet{liu2018lexicon} proposed a similar approach, but their strict keyword-matching approach could potentially miss implicit offensive language and informal expressions. 

However, it is important to note that most of the lexicon-based detection models are impacted by the dynamic nature of Chinese, with new words and phrases emerging rapidly. These models may fail in detecting hate speech that can take advantage of the language evolution to escape detection. Also, these models contribute to the propagation of subversive expressions in alternative forms, which can make the task of offensive language detection significantly more challenging. Nevertheless, despite its limitations, lexicon-based models remain widely used in the realm of Chinese offensive language.

\paragraph*{Machine Learning Based Models}
These models are built not only using supervised~\cite{jiang2021swsr} and semi-supervised~\cite{wu2019semi} approaches but also exploring the use of adversarial learning to solve this task~\cite{liu2020ai}. Most of the supervised approaches such as those presented by \citet{yuan2019weibohate}, provide a more adapted approach to identifying offensive language in a text. However, they too have their limitations. For example,\citet{yuan2019weibohate}'s model trained on WeiboHate, can suffer from imbalanced class distribution and the variability of annotator judgments, potentially impacting the quality of the training data. Several other models were presented in the literature that made use of ML techniques for classifying Chinese text as offensive or not.

\paragraph*{knowledge-based Models} These models, like the one discussed by \citet{liu2020combating}, seek to combat negative stereotypes and implicit biases in language. While this approach has potential, obtaining comprehensive common sense knowledge can be challenging. The success of such models depends on the availability and quality of the knowledge they are built upon, which can sometimes be incomplete or biased in itself, leading to limitations in performance.

\paragraph*{Multimodal Approaches}
\citet{zhang2020hurtmeplenty}'s study on HurtMePlenty showcases the use of multimodal approaches for nuanced hate speech detection in Chinese. Nevertheless, this approach faces challenges related to class imbalance within the dataset and the potential for platform-specific biases to affect the findings. Multimodal approaches, while promising, need careful consideration to account for these limitations and ensure robust results.

\paragraph*{Pretrained Language Models}
In Chinese offensive language detection, pre-trained models such as BERT\cite{devlin-etal-2019-bert} are commonly used in a supervised learning framework. In this approach, the model is initially trained on a large dataset containing labeled examples of offensive and non-offensive language. The training procedure involves fine-tuning the pre-trained language models using the task-specific dataset, which helps the model learn the specific features and characteristics of the Chinese language. Once this fine-tuning process is complete, the BERT-based model is used to classify new text as offensive or non-offensive with a high degree of accuracy. bert-base-chinese\footnote{\url{https://huggingface.co/bert-base-chinese}}  and roberta-base-chinese\footnote{\url{https://huggingface.co/roberta-base-chinese}} are the most commonly used ones for Chinese benchmarks~\cite{deng-etal-2022-cold}. 

\paragraph*{Cross-Cultural Transfer Learning}
Work from \citet{2303.17927} explores the impact of offensive language data from different cultural backgrounds on Chinese offensive language detection. It shows that Language Models are sensitive to cultural differences in offensive language, learning culture-specific biases that negatively impact their transfer ability. Additionally, the model they build quickly adapts to the target culture in the few-shot scenario, even with very limited Chinese examples. Such findings suggest that there exists a need to address the cross-cultural aspects of offensive language detection.

The experiments of translating Chinese and Korean datasets(COLD and KOLD respectively) into English showed similar trends, demonstrating that cross-cultural transfer is not only due to linguistic similarities. The data also displays experimental results indicating that the correlation between the ability to detect offensive language and target cultural knowledge follows a pattern similar to that of an increasing logarithmic function. These findings offer promising opportunities for low-resource offensive language detection systems.

\paragraph*{User Behavior Encoding}
\citet{8757714} showed that taking user information such as past comments and number of followers into account can significantly increase the accuracy of offensive language detection, especially when considering recall and F1 score metrics. This work was the first to propose an estimation of the potential aggressive score of a user and dynamically update this score. The model is trained using a corpus compiled from the Douban Chinese film reviews \cite{douban_corp}.\footnote{Douban (\url{https://douban.com/})is a Chinese movie recommendation website like IMDB where users can rate each movie with their comments.}
The authors suggested updating their work in the future with more data and SOTA techniques, such as mocking detection and larger neural networks that consider more user details, such as interpersonal relationships, hobbies, and interests. However, such a method might raise ethical considerations such as violating people's privacy and providing negative labels that would cause discrimination.


\section{Research Gaps in Chinese Offensive Language Detection}
\label{sec:potential}

\begin{table*}
\centering
\begin{tabular}{l|l|p{15em}}
\hline
\textbf{Challenges} & \textbf{Example} & \textbf{Explanation} \\
\hline
Mislabeled data & Women often get paid less & Facts are often considered offensive \\
Variety of cultural context & 咸家富贵 & The phrase means differently under different contexts \\
Subversive expressions & 草泥马/操你妈 & Such phrases can be undetectable \\
\hline
\end{tabular}
\caption{Examples of Challenges in Detecting Offensive Language in Chinese}
\label{table:challenges}
\end{table*}

Despite the decent amount of research on Chinese offensive language detection, there are still several important research gaps that need to be addressed in order to further advance the performance of the systems.

\paragraph*{\textbf{Context-aware Offensive Language Detection}}
Most of the existing models for offensive language detection only consider the
text of a comment, tweet, or post without taking into account its broader
context. In many cases, the same words or phrases can be used in a neutral or even positive way, or in an offensive way, depending on the surrounding words and the topic being discussed. This can lead to false positives and false negatives in offensive
language detection.

For example, "业务水平太高了！牛逼！"(The skills (possessed by someone) is so awesome) is not offensive if we neglect the context from where it exists. However, if such a comment appears under a news title that depicts some failures, such as a medical accident, the phrase mentioned above can be offensive to the doctor. Thus, textual context is important in detecting offensive language in Chinese as they often contain essential information. Therefore, there is a growing interest in developing approaches that take into account the textual context of the comments or tweets. One of the potential solutions is to incorporate topic modeling techniques to identify the topic of the comment or tweet and use this information as an additional input to the offensive language detection model.

This would allow the model to better understand the intended meaning of the comment and distinguish between offensive and non-offensive uses of certain words or phrases. Such models could be particularly useful in cases where the topic itself may be inflammatory or controversial, which can affect the interpretation of the comment text. Addressing this research gap could help improve the accuracy and effectiveness
of offensive language detection systems in Chinese social media. 

\paragraph*{\textbf{Limited offensive language variety and unclear severity boundaries}}

Offensive language is a complex and multi-dimensional phenomenon that can take many forms, ranging from direct hate speech and discriminatory language to more subtle forms of sarcasm. In the Chinese language context, there is a lack of unified research on the varieties of offensive language types and the clear boundaries of the level of offensive language, including hate speech, sarcasm, adult content, and cyberbullying. Additionally, there is a need to establish clear boundaries between biased attitudes, acts of bias, discrimination, and bias-motivated violence.

Despite the increasing awareness of the negative effects of offensive language, current research on Chinese offensive language detection tends to focus on a narrow range of categories and fails to capture the full complexity of this phenomenon. For example, some studies have focused solely on sexist language~\cite{jiang2021swsr}. The lack of differentiation may lead to ineffective detection and prevention strategies, as different types of offensive language require
different approaches.

Furthermore, there still exists a research gap in terms of defining clear boundaries of the level of offensiveness. While some studies have attempted to address the issue of detecting hate speech and discriminatory language, there lacks a clear agreement and distinction on what constitutes biased attitudes and behaviors in different contexts. This creates additional 
challenges to developing effective and trustworthy models for detecting offensive languages that are based on the complexity of the Chinese language and culture. Thus, there exists a need for a more comprehensive dataset that includes a broader range of offensive language, including various levels of biased attitudes and behaviors, to support the development of accurate and contextually relevant detection models. 

\section{Offensive Language Detection in Chinese: Challenges}
\label{sec:challenges}
Offensive Language Detection presents several challenges that must be addressed in order to overcome potential issues and achieve meaningful and effective results. Table~\ref{table:challenges} presents a few examples of these challenges.
 
\subsection{Data Labeling Problems and Subjectivity}
One of the major challenges in Chinese offensive language detection is the existence of mislabelled data, which can result from various factors such as the contested definition of offensiveness, which can vary depending on cultural and social context. This can lead to inconsistent annotations of offensive language in different datasets, as well as disagreement among annotators. 

Another factor that can contribute to obtaining mislabeled data is annotator subjectivity. Annotators tend to misunderstand the nature and scope of offensive language. In fact, annotators may have different levels of sensitivity and tolerance towards certain types of language and may interpret the same text differently based on their personal beliefs and values. This can lead to inconsistencies in labeling and may result in a lower recall rate and precision. Hence, spending time on training and educating annotators about the subject and providing clear guidelines and definitions for labeling offensive language is crucial in obtaining accurate and reliable data. Additionally, regular quality checks and measures to ensure an inter-annotator agreement can also help mitigate the impact of potential misunderstandings and inconsistencies. 

\subsection{Cultural Context Variety}
One of the biggest challenges in Chinese offensive language detection is accounting for the variety of cultural contexts that exist within the language. Due to the vastness and diversity of Chinese culture, language use can vary widely depending on the context and the cultural background of the speaker or writer. This presents a significant challenge for Chinese offensive language detection models, as what might be considered offensive in one cultural context may not be considered offensive in another. As the example mentioned above, the phrase ``ham ga fo gui '''(咸家富贵/全家富贵) means
the same thing as the phrase ''ham ga can''(咸家铲/全家铲), where the prior one in a Mandarin context being ``hope your family gets rich'', and the later being ``hope you whole family dies''. The phrase ''fo gui'', in such context refers to the paper money that dead people ``receive'' when they are dead, rather than a sincere wish for the individual to become rich. Such a phrase in this particular context acts as a way of sarcasm, rather than a sincere compliment, which adds another layer of complexity to detecting offensive language
in Chinese. To address this challenge, offensive language detection models must take into account the cultural context in which the language is being used. This requires a deep understanding of Chinese culture and language use, as well as the ability to identify and analyze the cultural factors that influence language use. A qualitative analysis of the social media content matters could be suggested.

\subsection{Neologism in Chinese}
Neologism presents a unique challenge for detecting offensive language in Chinese. These are linguistic expressions, words, or phrases that are often used to circumvent censorship or to express taboo or sensitive topics in a covert manner \cite{Wiener2011GrassMudHT}. Even worse, people would embed implicit offense inside a seemingly polite and acceptable phrase.  These phrases are often deeply embedded in specific cultural contexts and may be difficult to detect without an understanding of the relevant cultural references and language nuances. An example of subversive expressions from a post in Baidu Tieba, a popular Chinese online forum platform where users can discuss various topics and share information, is given in Figure~\ref{figure:tieba}.\footnote{\url{https://tieba.baidu.com/}}
\begin{figure}
    \centering
    \includegraphics[width=0.9\linewidth]{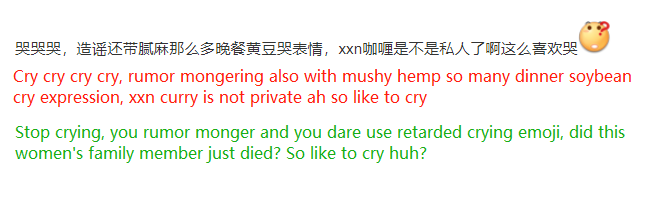}
    \caption{An example of the use of subversive phrases on Tieba}
    \label{fig:tieba}
\end{figure}

This kind of alternative expressions can not be detected by standard offensive language detection systems, which rely on detecting specific keywords or patterns of language. Therefore, it is crucial to define a model that takes into account the cultural context and the use of alternative symbolic infrastructures when developing offensive language detection algorithms in Chinese. However, to the best of our knowledge, there is no research work that explores how potentially harmful neologism can be detected using existing NLP technology. Tackling the issue of harmful Neologism can help preserve the original meaning of words and lessen the effect of cyberbullying and thus be beneficial to the development of the Chinese language.

\section{Addressing the current issue}
\label{sec:address} 
To address the challenges in detecting offensive language in Chinese text, several strategies can be employed:
(1) Incorporate context-aware detection using topic modeling and discourse-level analysis, and consider dialogue models for contextual cues; (2) Develop more diverse datasets covering various offensive language types, including sarcasm and regional variations; (3) Enhance annotation quality with clear guidelines, hierarchical classification, and better annotator training; (4) Capture cultural references and creative expressions in knowledge bases for covert offensive language detection; (5) Train models jointly on sarcasm and offensive language datasets to recognize connections; (6) Generate realistic adversarial examples and employ semi-supervised approaches for model resilience; (7) Conduct cross-geography transfer learning studies to adapt models to different cultures while avoiding bias with techniques like gradient reversal; (8) Foster collaboration through shared offensive language resources and collaborative annotation projects; (9) Engage users and utilize participatory design approaches to improve model cultural awareness.

\section{Conclusion}
\label{sec:conclusion}
In this paper, we explored the current benchmark and approaches for offensive language detection for Chinese and presented the models and tools built to address this task. We also identified the challenges and research gaps for this task posed mainly by the inherently complex nature of the Chinese language. Our study highlights the need to address the unique difficulties due to cultural ambiguity. Future research should consider cultural context when detecting offensive language, and explore ways to detect subversive expressions and sarcasm. 
\bibliography{test,anth}
\bibliographystyle{Frontiers-Vancouver}
\end{CJK*}
\end{document}